# Efficient Decision-Theoretic Planning: Techniques and Empirical Analysis

**Peter Haddawy** **AnHai Doan**
Department of EE&CS
University of Wisconsin-Milwaukee
Milwaukee, WI 53201
{*anhai, haddawy*}*@cs.uwm.edu*

**Richard Goodwin**
School of Computer Science
Carnegie Mellon University
5000 Forbes Ave. Pittsburgh, PA 15213
*rich@cs.cmu.edu*

## Abstract

This paper discusses techniques for performing efficient decision-theoretic planning. We give an overview of the DRIPS decision-theoretic refinement planning system, which uses abstraction to efficiently identify optimal plans. We present techniques for automatically generating search control information, which can significantly improve the planner's performance. We evaluate the efficiency of DRIPS both with and without the search control rules on a complex medical planning problem and compare its performance to that of a branch-and-bound decision tree algorithm.

## 1 Introduction

In the framework of decision-theoretic planning, uncertainty in the state of the world and in the effects of actions are represented with probabilities; and the planner's goals, as well as tradeoffs among them, are represented with a utility function over outcomes. Given this representation, the objective is to find an optimal or near optimal plan. Finding the optimal plan requires comparing the expected utilities of all possible plans. Doing this explicitly is computationally prohibitive in all but the smallest of domains. This is due to the large space of possible plans that must be searched and to the fact that probabilistic plan evaluation entails high computational cost.

Researchers have taken various approaches to dealing with this complexity. One approach has been to focus on solving part of the problem by working with probabilities and categorical goals [10, 6] or by planning with goal-directed utility functions but under complete certainty [13]. These approaches are able to gain some efficiency by exploiting the structure that arises due to the use of categorical goals, deterministic actions, or restrictions on the form of the utility function.

A second popular approach has been to work with a constrained and highly structured problem representation, exemplified by the discrete Markov process-based planners [2, 1]. The model assumes a finite state space and a limited class of utility functions. Even so, existing algorithms for both completely and partially observable Markov processes have exponential running time in terms of the number of domain attributes and are thus applicable for only small domains.

A third approach uses qualitative techniques to filter out classes of obviously bad plans, thus avoiding costly plan evaluation [11]. While such qualitative dominance proving can be highly efficient, it requires much structure and is typically not powerful enough to identify the optimal plan. At some point one must resort to quantitative reasoning about expected utility in order to evaluate tradeoffs.

In large domains we expect that even if qualitative techniques are used as a filter, the remaining space of possible plans will be too large to exhaustively examine. To search such a space effectively, we have developed the DRIPS decision-theoretic refinement planning system which obtains its efficiency by exploiting information provided in an abstraction hierarchy. The ability to structure actions into an abstraction hierarchy requires the domain to contain actions that can be grouped according to similarity but imposes no other requirements concerning the structure of the domain or the utility function. By using abstraction, the planner can eliminate suboptimal classes of plans without explicitly examining all plans in the class.

A decision-theoretic planning problem can be characterized in terms of a number of parameters. We have a set of states of the world, which may not be completely observable, a set of actions from which plans can be constructed, a class of utility functions for which we can plan, and some time horizon over which we will consider plans. In this paper we allow the state space to be infinite. States are only partially observable. We assume a finite set of actions and plans. A plan is a sequence of actions. We do not restrict the allowable forms of the utility function. The planning horizon is assumed to be finite. Given this framework, we are interested in finding the optimal plan.

The rest of this paper is organized as follows. We first present the representational framework used throughout the paper. We then present an overview of the DRIPS planner and of the abstraction theory used by the planner to prune the space of plans. In the next section we present two methods for efficiently controlling the plan elaboration and pruning process. We then empirically evaluate the efficiency of DRIPS by applying it to the medical planning problem of selecting the optimal test/treat strategy for managing patients suspected of having deep-vein thrombosis (DVT) of the lower extremities. We show that DRIPS significantly outperforms a standard branch-and-bound decision tree evaluation algorithm on this domain and compare the efficiency of the planner with and without the search control. We finish with a discussion of future research.

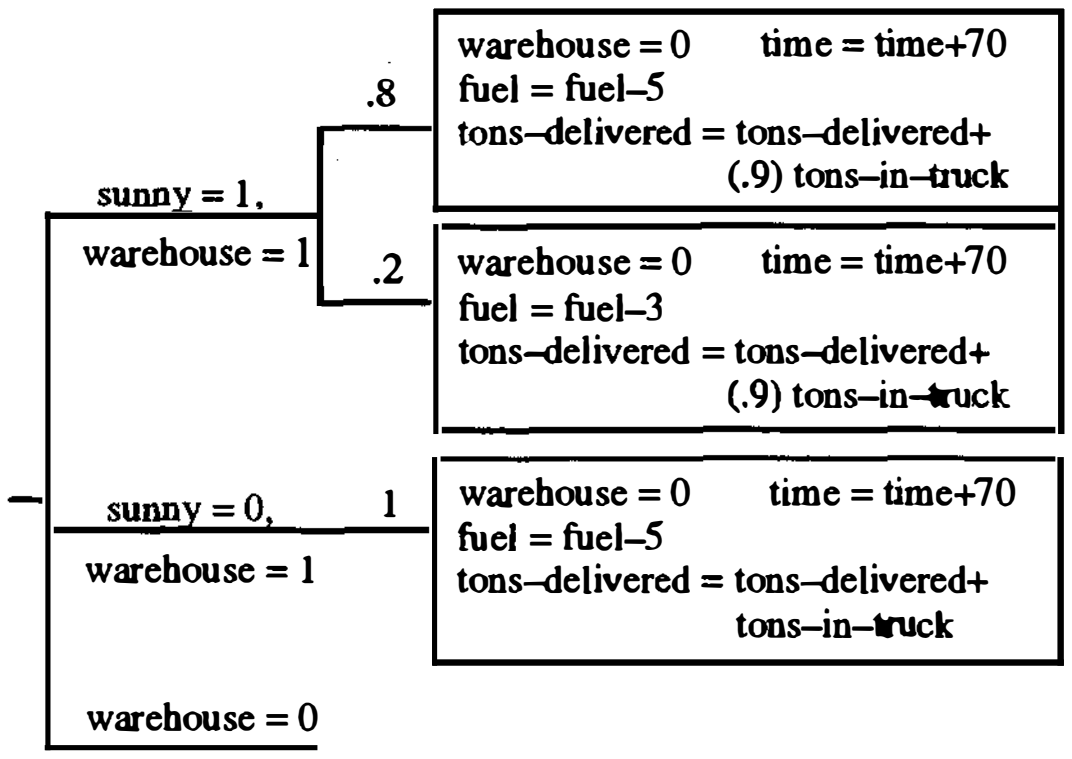

Figure 1: Action "Deliver-tomato".

## 2 Representation

**World Model** We describe the world in terms of chronicles, where a chronicle is a complete specification of the world throughout time. We take time to be continuous and we describe chronicles by specifying the values of discrete and continuous attributes at various times, for example fuel($t_0$) = 10. We express uncertainty concerning the state of the world with a set of probability distributions over chronicles. We express such a set by assigning probability intervals to (attribute, value) pairs at various times.

**Action Model** An action is represented by a finite set of tuples $\langle c_i, p_i, e_i \rangle$ called branches, where the $c_i$ are a set of mutually exclusive and exhaustive conditions, the $p_i$ are probabilities, and the $e_i$ are effects. The intended meaning of an action is that if the condition $c_i$ is satisfied at the beginning of the action then with probability $p_i$ the effect $e_i$ will be realized immediately following the action. This representation form is used in [10] and utilized by work in Markov Decision Process [2, 1]. In that work an action condition or effect is specified by a set of *propositional assignments*, such as *painted* $\wedge \neg$ *hold-block*. We extend the representation by also allowing *metric assignments* in action conditions and effects, such as $fuel(t_2) = fuel(t_1) - 5$; we further allow branch probabilities $p_i$ to be represented by intervals instead of single numeric points. These relaxations substantially enhance the expressiveness of the representation. We assume that changes to the world are limited to those effects explicitly described in the agent's action descriptions.

Figure 1 shows an example of an action description. The action Deliver-tomato describes the activity of delivering tomatoes from a warehouse to the market. The action is conditioned on the weather and the start location. The first tuple, for example, says that if the weather is sunny and the start location is at the warehouse ($sunny = 1, warehouse = 1$) then with the probability .8 the action takes 70 units of time ($time = time + 70$), consumes 5 units of fuel ($fuel = fuel - 5$), 10% of the tomatoes in the truck are spoiled because of the sun ($ton - delivered = ton - delivered + .9 * ton - intruck$), and after the action the location of the truck is not at the warehouse ($warehouse = 0$). Notice that propositional assignments such as $warehouse = 0$ are represented in a format identical with that of metric assignments.

## 3 Decision-Theoretic Refinement Planning

### 3.1 Abstracting Actions

The DRIPS planner primarily uses two types of abstraction: interaction-abstraction and sequential abstraction. The idea behind inter-action abstraction is to group together a set of analogous actions. The set is characterized by the features common to all the actions in the set. We then can plan with the abstract action and infer properties of a plan involving any of its instances. Formally, an inter-action abstraction of a set of actions $\{a^1, a^2, ...a^n\}$ is an action that represents the disjunction of the actions in the set. The actions in the set are called the *instantiations* of the abstract action and are considered to be alternative ways of realizing the abstract action. Thus the $a^i$ are assumed to be mutually exclusive.

To create an inter-abstraction of a set of actions $\{a_1, a_2, ..., a_n\}$ we do the following. Group the branches of the action descriptions into disjoint sets such that each set contains at most one branch from each action description. For each set $s$ that contains fewer than $n$ branches, add $n - |s|$ branches, each with the effect of one of the branches already in the set and with condition False and probability zero. The effect of an abstract branch is any sentence entailed by each of the effects of the branches in the set. The condition is the disjunction of the conditions on the branches in the set. The probability is specified as a range: the minimum of the probabilities of the branches in the set and the maximum of the probabilities of the branches in the set.

A sequential abstraction is essentially a macro operator that specifies the end effects of a sequence of actions, as well as the initial conditions under which those effects are achieved, without specifying changes that occur as intermediate steps due to the individual actions within the sequence. Thus the information about the state of the world during the execution of the sequence of actions is abstracted away. We abstract an action sequence $a_1a_2$ by pairing every branch of $a_1$ with every branch of $a_2$ and create an abstract branch for each pairing. The condition on the abstract branch is the conjunction of the conditions on the paired branches; the probability is the product of the probabilities on the paired branches; and the effect is the composition of the effects.

We have implemented tools that automatically create inter-action abstractions [5] and sequential abstractions [3]. For a general theory of action abstraction which includes intra-action and sequential abstraction see [4].

### 3.2 The DRIPS Planner

A planning problem is described in terms of an initial state distribution, a set of action descriptions, and a utility function. The space of possible plans is described by an abstraction/decomposition network, supplied by the user. An abstract action has one or more sub-actions, which themselves may be abstractions or primitive actions. A decomposable action has a subplan that must be executed in sequence. The description of the abstract actions are created by inter-action abstraction and the descriptions of the decomposable actions are created by sequential abstraction. An example network is shown in Figure 3. A plan is simply a sequence of actions obtained from the net. The planning task is to find the sequence of actions for those represented in the network that maximizes expected utility relative to the given probability and utility models.

DRIPS finds the optimal plan by building abstract plans, comparing them, and refining only those that might yield the optimal plan. It begins with a set of abstract plans at the highest abstraction level, and subsequently refines the plans from more general to more specific. Since projecting abstract plans results in inferring probability intervals and attribute ranges, an abstract plan is assigned an expected utility interval, which includes the expected utilities of all possible instances of that abstract plan. An abstract plan can be eliminated if the upper bound of its expected utility interval is lower than the lower bound of the expected utility interval for any other plan. Eliminating an abstract plan eliminates all its possible instantiations. When abstract plans have overlapping expected utility intervals, the planner refines one of the plans by instantiating one of its actions. Successively instantiating abstract plans will narrow the range of expected utility and allow more plans to be pruned.

Given the abstraction/decomposition network, we evaluate plans at the abstract level, eliminate suboptimal plans, and refine remaining candidate plans further until only optimal plans remain. The algorithm works as follows.

1. Create a plan consisting of the single top-level action and put it into the set plans.
2. Until there is no abstract plan left in plans,
   - Choose an abstract plan P. Refine P by replacing an abstract action in P with all its instantiations, or its decomposition, creating a set of lower level plans $\{P_1, P_2, ..., P_n\}$.
   - Compute the expected utility of all newly created plans.
   - Remove P from plans and add $\{P_1, P_2, ..., P_n\}$.
   - Eliminate suboptimal plans in plans.
3. Return plans as the set of optimal plans.

Since DRIPS only eliminates plans that it can prove are suboptimal and if run to completion it explores the entire space of possible plans, it is guaranteed to find the optimal plan or plans. Notice that the algorithm can be stopped at any time to yield the current set of candidate plans. This feature can be exploited to flexibly respond to time constraints.

## 4 Control Strategies

The run time efficiency of the DRIPS planner depends on effectively controlling the search through the space of abstract plans. The DRIPS algorithm contains two non-deterministic choice points in its second step. The first choice is to select a plan from the set of abstract plans with overlapping expected utility. The second choice is to select an abstract action within the plan for expansion.

Consider selection of the plan to be refined. There are two ways a potentially optimal plan can be eliminated from consideration: either the upper bound of the plan is lowered below the highest lower bound or the highest lower bound is raised above the level of the upper bound of the plan. Notice that since the abstract plan with the maximal upper bound on expected utility may contain an optimal primitive plan, that abstract plan must be expanded to insure that we have found the complete set of optimal plans. So at any point in the search the abstract plan with the current maximal upper bound will eventually need to be expanded. Thus the strategy of always expanding a non-primitive plan with the maximum upper bound on expected utility will lead to the optimal solution length.

The selection of an action to expand is more problematic. Selecting actions that when expanded produce plans with greater reductions in the range of expected utility facilitates pruning and leads to more efficient planning. In this section we present and discuss three approaches to selecting actions for expansion. The first approach uses a simple heuristic, the second uses supplied priorities, and the third uses sensitivity analysis.

The default action selection method uses a simple heuristic that chooses the first abstract action in the plan for expansion, since picking a specific initial action can help to differentiate the set of abstract plans. This method is domain independent, and requires no additional effort on the part of the domain designer. In practice, we have found that the heuristic performs significantly better than random action selection. However, the first action heuristic is not a well informed heuristic and cannot take advantage of the structure of the domain to guide the search. For the purposes of this paper, we use this heuristic as a strawman to provide a baseline against which to compare the performance of the other two action selection methods.

A fixed priority action selection method allows the domain designer to assign a priority to each abstract action. At each choice point, the action with the highest priority is selected for expansion. Ties are broken by selecting the first action with the highest priority. This method allows the domain designer to encode extra information about the domain that can server as a better informed heuristic. Like the default method, it is cheap to compute. The disadvantage of this method is that it places an extra burden on the domain designer to set the priorities. One method we have used to set priorities is to assign low priority to actions that have similar instantiations and higher priority to actions with dissimilar instantiations. Preferentially refining actions with dissimilar instantiations should lead to less overlap in the expected utility of the subplans. We observe the performance of the planner on example problems to validate the priority assignments.

### 4.1 Sensitivity Analysis

The third method of selecting actions for expansion uses a sensitivity analysis to select actions to which the bounds on expected utility are most sensitive. The sensitivity analysis uses the structure of the actions in the domain and the utility function to adapt the search control to the specific abstract plan to be expanded. Tailoring the search control in this way produces a more informed heuristic and should lead to less search. The disadvantage of this method is that the sensitivity analysis involves some computational overhead. The added cost must be traded off against any reduction in the amount of search needed.

The sensitivity analysis used is structural, based on the method used to select plans for expansion. Plans selected for expansion are those with the highest bound on expected utility. If an action expansion creates subplans with the same upper bound on expected utility, then the subplans will be immediately selected for further expansion. Preferentially expanding actions that can cause larger changes in the upper bound of expected utility should reduce the likelihood that the sub-plans will be chosen for further expansion. This should lead to increased search efficiency.

In the rest of this section, we present sensitivity analysis for a general utility model proposed [7]. The expected utility of a plan $p$ is the sum of the utilities of the possible chronicles weighted by the probability of each chronicle $\mathbf{EU}(p) = \sum_{c \in \{chronicles\}} U(c) \cdot P(c)$. Utility of each chronicle, $U(c)$, is the weighted sum of the utility of goal satisfaction UG and residual utility UR, $U(c) = UG(c) + k_r UR(c)$.

The utility functions, UG and UR, form part of the problem description input to the planner. The sensitivity analysis requires two additional functions that give the possible change in the upper bound of the utility functions as a result of expanding an action. The $\Delta UG^{+}(chronicle, action, plan)$ function returns the maximum change in the upper bound on utility of goal achievement for a chronicle if the given action in the given plan is expanded. A second function, $\Delta UR^{+}(chronicle, action, plan)$, similarly returns the maximum change in the upper bound on the residual utility. The $\Delta UG^{+}$ and $\Delta UR^{+}$ functions can be derived from the UG and UR functions respectively. However, since the UG and UR functions can be arbitrarily complex, the two additional functions must be supplied by the domain designer.

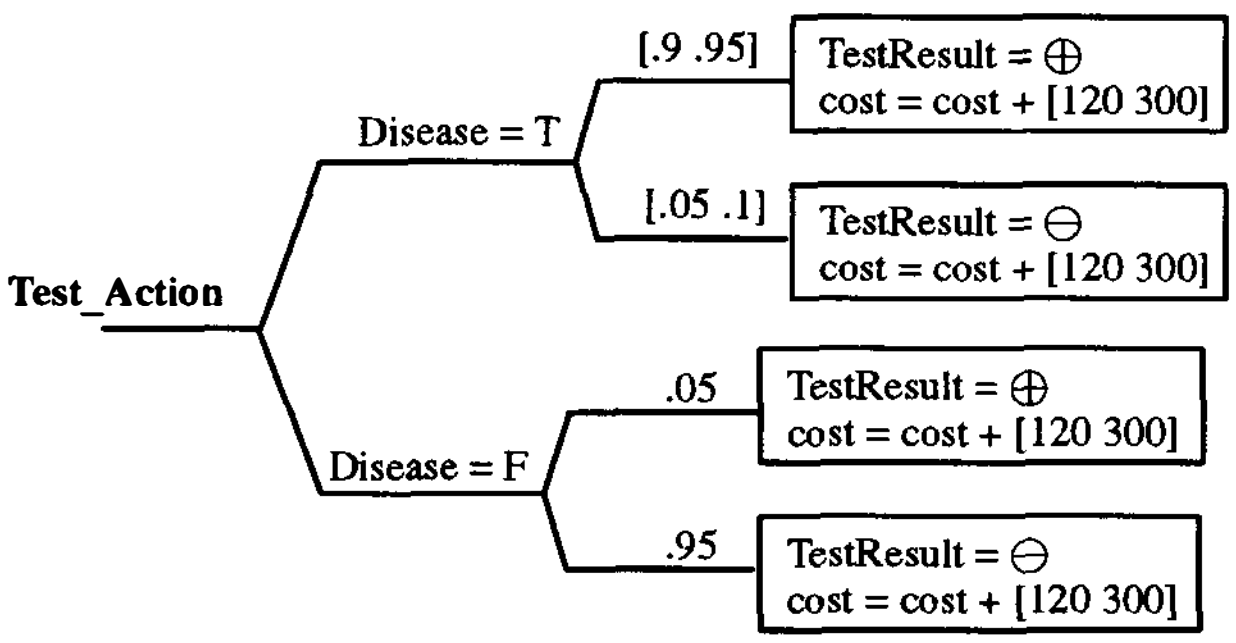

Figure 2: Abstract test action representing tests with different costs and different false negative probabilities.

To get an idea of what is involved in creating the $\Delta UG^{+}$ and $\Delta UR^{+}$ functions, consider the abstract test action in figure 2. The abstract test action has several possible instantiations that account for the range in the cost and the range in the probability of a false negative. Further suppose that the UR function is: $UR(c) = -(cost + COST_FATALITY)$ if the patient dies and $UR(c) = -(cost)$ otherwise. The chronicle passed to the $\Delta UR^{+}$ can be used to determine which branch of the action the chronicle corresponds to. The assignments of cost and TestResult in that branch give the effect that the action can have on the plan. In this example, the possible change in the UR value corre-sponds to the difference in the costs of the tests ($300-$120) = $180.

A third function, $\triangle P(chronicle, action, plan)$, returns the maximum change in the upper and lower bound of the probability of a chronicle that can result from expanding the given action. This function depends only on the structure of the actions and not on the form of the utility function. It is implemented domain independently in the planner. In the medical example above, refining the test action could reduce the range in the probability of a false negative.

The overall sensitivity of the upper bound on expected utility can be calculated by combining the $\triangle$ functions. The following equation gives the least upper bound on utility

$$\hat{U}(c,a,p) = (UG^{+}(c) - \triangle UG^{+}(c,a,p)) + K_r \cdot (UR^{+}(c) - \triangle UR^{+}(c)),$$

where $UG^{+}$ and $UR^{+}$ represent the current upper bounds on $UG$ and $UR$, respectively. Calculating the effect on probability is more complicated. The upper bound on expected utility can be lowered either by decreasing upper bound on the probability of high utility chronicles or increasing the lower bound on the probability of low utility chronicles. Let the function $\hat{P}$ give the probability of each chronicle, with the probability bounds adjusted by up to $\triangle P$ and subject to the condition that the sum of the probabilities be 1. The overall sensitivity is then the upper bound on expected utility minus the least upper bound on the expected utility after action expansion,

$$sensitivity(a,p) = EU^{+} - \min\left\{ ub\left(\sum_{c} \hat{U}(c,a,p)\cdot \hat{P}(c,a,p)\right)\right\}.$$

When selecting an action for expansion, the computation needed to expand the action and evaluate the subplans needs to be taken into account. The final weighting for each action is the ratio of the sensitivity divided by the cost of expanding the action. For simplicity, we estimate the cost by counting the number of sub-plans that would be have to be evaluated, which is equal to the number of instantiations of the action. The action with the highest weighting is then selected for expansion.

# 5 Empirical Analysis

## 5.1 Comparing DRIPS with Branch and Bound Algorithms

Appropriate management of patients with suspected acute deep venous thrombosis (DVT) of the lower extremities is an important and complex clinical problem. To evaluate the effectiveness of DRIPS, we constructed a model for diagnosis and treatment of DVT[1], based on data from an article that compared various different management strategies [8]. To encompass all of the strategies described in the original model, our model incorporated up to four tests, with a maximum of three 7-day waiting periods between tests. The test procedures included contrast venography (Veno) and two non-invasive tests (NIT): impedance plethysmography (IPG) and real-time ultrasonography (RUS). Treatment, which consisted solely of anticoagulation therapy, included unconditional actions (e.g., Treat All) and conditional actions (e.g., Treat if thigh DVT seen on venography [Treat if Veno Thigh+]). The utility function used for the analysis was defined as the sum of the costs of tests and treatment and the costs associated with the state of the patient at the end of the plan.

A portion of the abstraction/decomposition network for the DVT domain is shown in Figure 3[2]. The most abstract action, Manage DVT, is an abstraction of six actions: No_Tests_and_Treat, Veno_Tests, NIT_Tests, Two_Tests, Three_Tests, Four_Tests. (The number of tests represents the length of the longest allowed sequence of tests.) Each of these actions further decomposes into a sequence of actions. For example, NIT_Tests decomposes into NIT, Treat_NIT. NIT can be instantiated as IPG or RUS. Our model for management of suspected DVT encompassed 6,206 concrete plans; for example, one complete plan (an instance of the Two_Tests action) is "IPG, Wait_7d_if_NIT-, Veno_if_NIT-, Treat_if_Veno_Any+."

We ran DRIPS with several variations of the utility function and in all cases it successfully identified the optimal plan. This was verified by comparison with a decision-tree evaluation algorithm. The results produced by DRIPS differed from those reported in the reference manuscript [8]. In reviewing these results, we discovered that DRIPS had uncovered an error in the original study [9].

In evaluating this model, DRIPS evaluated only 655 abstract and concrete plans out of a total of 6,206, yielding a pruning rate of 89%.[3] In order to demonstrate the efficiency of DRIPS in practice, we compared its performance on several variations of this problem to that of a standard branch-and-bound algorithm for evaluating decision trees. Figure 4.a shows the running time for DRIPS and the running time for the decision tree branch-and-bound algorithm at values of cost of fatality ranging from $50,000 to $500,000. DRIPS outperformed the branch-and-bound algorithm at all values. In the most extreme case, the running time of DRIPS was only 15% that of the branch-and-bound algorithm.

[1] The DRIPS code and the DVT domain are available via www at http://www.cs.uwm.edu/faculty/haddawy.

[2] It should be noted that the network structure followed naturally from our understanding of the problem and thus took very little time to produce, but producing the abstract action descriptions was a rather laborious task.

[3] On the clinical planning problem of finding the optimal test/treat strategy for diabetic patients suspected of having a foot infection, DRIPS evaluated only 13 out of 258 possible plans, achieving a pruning rate of 95%.

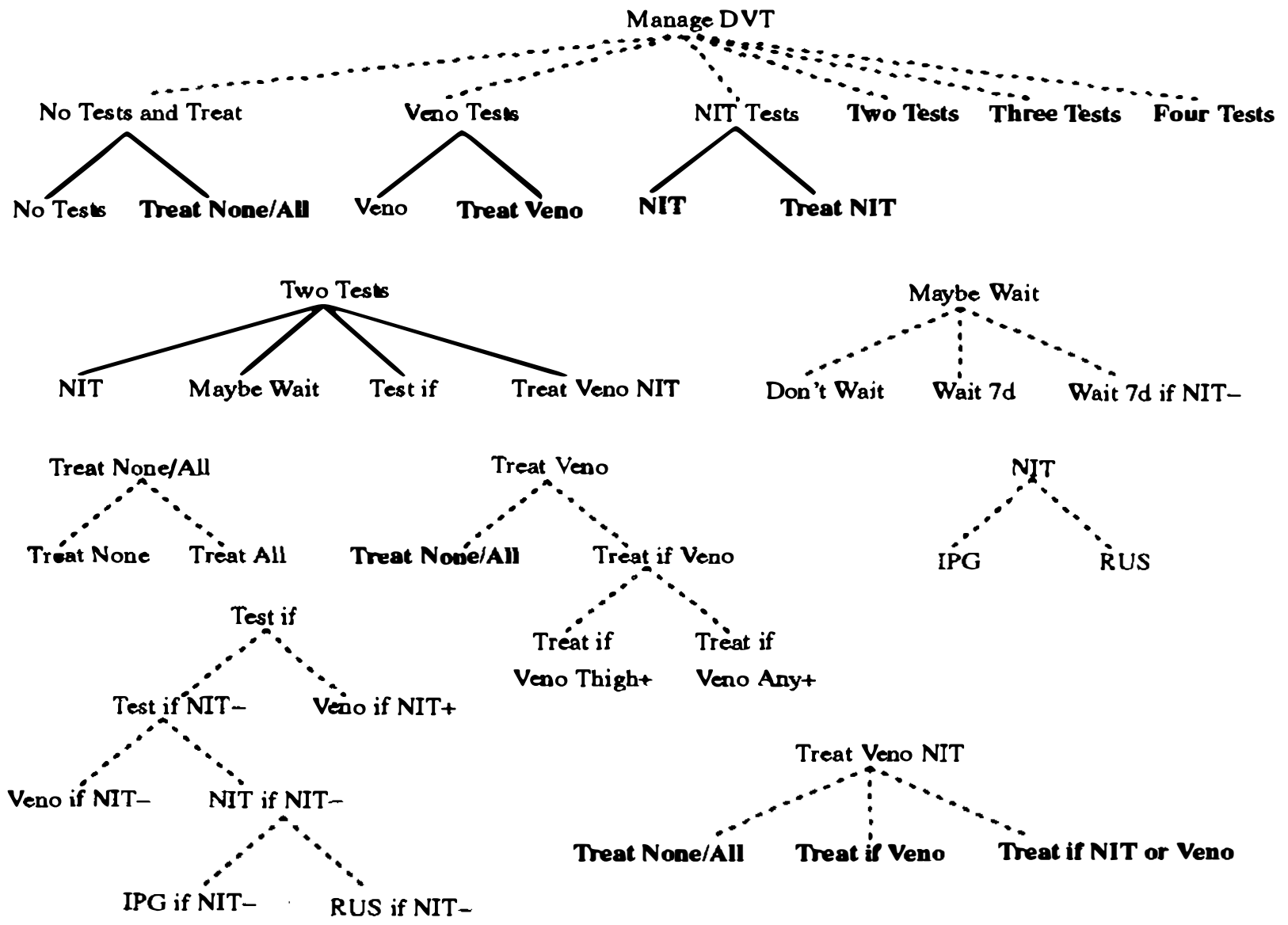

Figure 3: Abstraction/decomposition network. Abstraction relations are shown with dashed lines and decomposition relations are shown with solid lines. Actions shown in bold have decompositions or abstractions which are displayed elsewhere in the figure.

To examine how the efficiency of the two approaches varies as a function of problem size, we applied each approach to four versions of the DVT domain of increasing size. Figure 4.b shows the running times per plan for DRIPS and the branch-and-bound algorithm for each of the domains. Notice that the running time per plan for the branch-and-bound algorithm increases markedly as a function of problem size while the running time per plan for DRIPS actually decreases. This means that for this domain the DRIPS algorithm scales up much more effectively then the branch-and-bound algorithm. Figure 4.c shows that the memory usage of DRIPS also compares favorably to that of the branch-and-bound algorithm over this same suite of problems. In the most extreme case, DRIPS uses only 4.4% as much memory as the branch-and-bound algorithm. This can be explained by the fact that at any time DRIPS projects and evaluates only a small, constant number of plans, while evaluating the decision tree requires keeping track of all subtrees projected.

### 5.2 Comparing Control Strategies

The comparisons of the DRIPS algorithm with the decision tree algorithm in the previous section made use of the fixed priority control strategy. In this section, we repeat some of the tests to compare the different DRIPS control strategies. The results are shown in Figure 5.

For a small domain size, search control has little effect on efficiency. The results for the smallest DVT domain given in Figure 5.a show that all the strategies expand about the same number of plans and that running times are comparable. For this particular domain, the optimal strategy evaluates from 10 to 16 plans and all of the strategies perform nearly optimally.

In larger domains, search control becomes more critical. The results for the largest DVT domain, Figure 5.b, show a wide divergence in both the number of plans evaluated and in the running time. In the larger domain, pruning can significantly reduce the amount of search needed and effective search control leads to better pruning.

In larger versions of the DVT domain, the fixed priority control strategy does significantly better than the default first action heuristic. Fixed priority control adds little overhead and evaluates significantly fewer plans to produce a much improved running time. The performance of the strategy is significantly better for utility functions with a high cost of fatality, as is the default strategy. This can be partly accounted for by the fact that the domain designer only assigned priorities to some of the more significant abstract actions. In cases where priorities are equal or not assigned, the fixed priority strategy falls back to the default strategy. The first action heuristic does well for utility functions with a high cost of fatality because the value of better information increases since the cost of making a mistake is so high. The first actions in a treatment plan tend to be tests, and creating plans that use different tests can significantly differentiate their expected

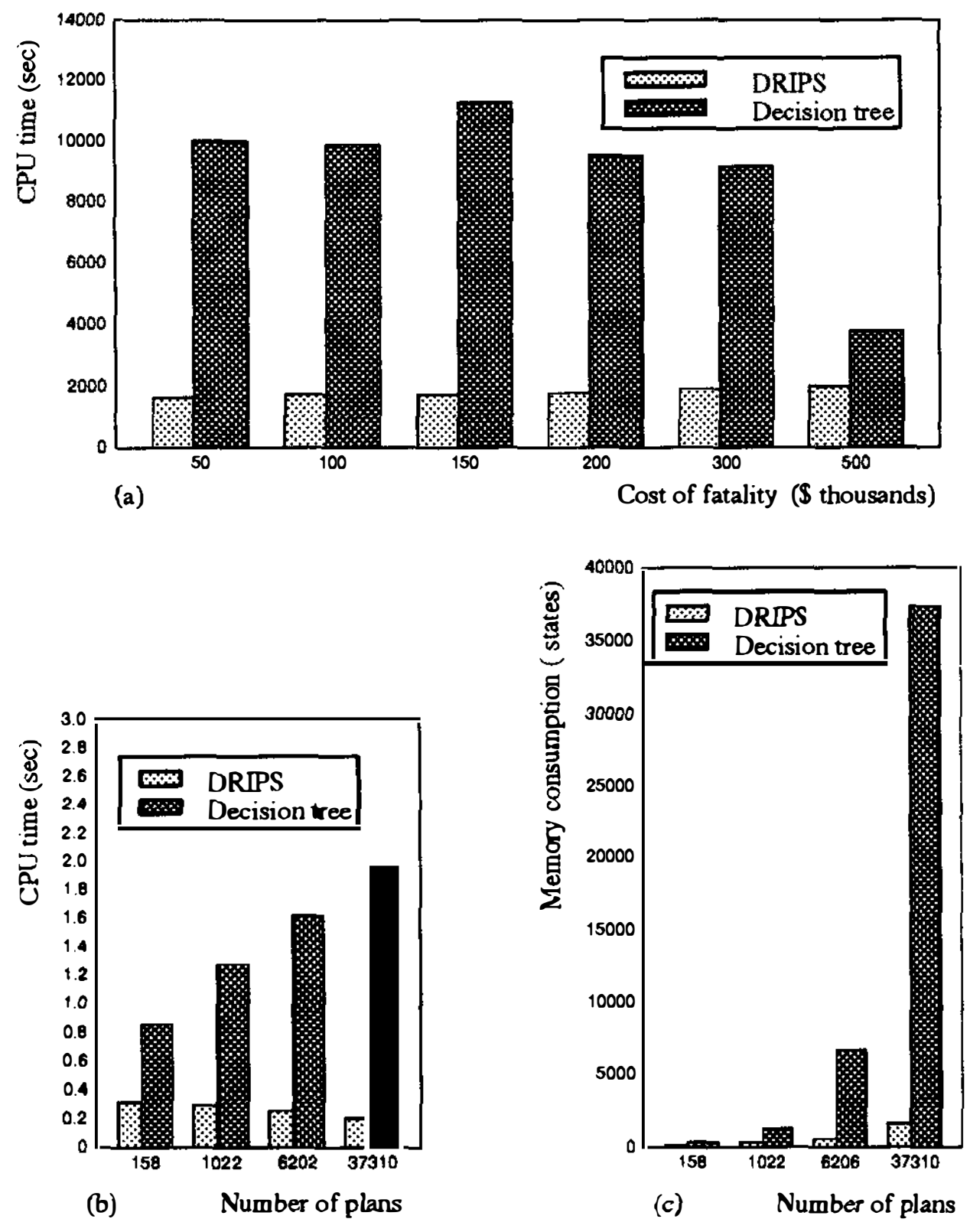

Figure 4: (a) Running times for DRIPS and a branch-and-bound decision tree evaluation algorithm for various costs of fatality. (b) Running times per plan and (c) memory consumption for DRIPS and the branch-and-bound algorithm for problems of increasing size. Memory consumption values represent the maximum number of world states that need to be stored at one time by each algorithm.

utilities and lead to better pruning.

The sensitivity analysis based control strategy also does significantly better than the default control strategy over the entire range of utility functions for large domains. The performance of this strategy is almost constant since the strategy adapts the search control to the changing utility function. As a result, the sensitivity analysis strategy does significantly better than the fixed priority strategy for lower costs of fatality, especially in terms of the number of plans evaluated. For higher costs of fatality, the fixed priority scheme only expands a few more plans and since it has a lower overhead, the running time is better. The running time of the sensitivity analysis strategy could be improved by optimizing the sensitivity analysis code or by doing only a partial sensitivity analysis. For example, the sensitivity could be calculated relative to some fraction of the most likely chronicles. The speedup would be linear in the inverse of the fraction of chronicles used, but this would have to be traded off against any degradation in the quality of search control.

# 6 Discussion and Future Research

The efficiency of our approach depends largely on how domain regularities are exploited to build the abstraction hierarchy. The applicability and the performance of the planner would be improved if methods could be devised to perform tight abstraction, and loss due to abstraction could be quantified. Devising such procedures and loss estimates has been shown to be possible for a limited class of domains [1]. In more complex domains with more expressive utility functions it is much harder to work out efficient abstraction procedures, although relatively good abstraction hierarchy can still be built by exploiting simple regularities in the domain and heuristics given by domain experts. If we know, for example, that two alternative actions differ only in the value of an attribute, say *cost*, and the change in the value of *cost* produces very little change in the value of the utility function, then the abstraction of the two actions will be tight. The sensitivity analysis presented in Section 4.1 exploites such simple domain and utility regularities to control plan elaboration. We are currently working on methods for automatically generating good abstractions for use by the DRIPS planner. The method starts by assigning weights to the domain attributes by analyzing the utility function and the primitive action descriptions. A clustering algorithm then uses the primitive action descriptions and weights to group together similar actions.

A significant contribution of our approach is the ability to perform decision theoretic planning in richer domains and utility models than those in the Markov approach. Since existing algorithms for the Markov models have exponential running time in the number of domain attributes, and do not exploit much of the domain regularities it would be interesting to compare our approach with these methods.

To further improve the efficiency of DRIPS, we are currently working on a technique that exploits stochastic dominance to eliminate suboptimal plans without computing their expected utilities. A relation called *stochastic dominance* can be established between the probability distributions of two random variables if the two distributions satisfy certain constraints [12]. Random variables representing domain attributes at different time points are typically related to one another by formulas via action effect assignments, and transformations caused by actions often create dominance situations, which can be verified fairly easily without knowing the exact probability distributions of the random variables. For any two plans $p_1$ and $p_2$ we can then try to locate two joint probability distributions $P$ and $Q$ over a subset of domain attributes such that $P$ and $Q$ are obtained at some timepoint during projecting $p_1$ and $p_2$, respectively, and $P$ dominates $Q$. The dominance of $P$ over $Q$ can be used to to prove that $p_1$ has a higher expected utility than $p_2$ (or vice versa) [12]; $p_2$ can then be eliminated from further consideration.

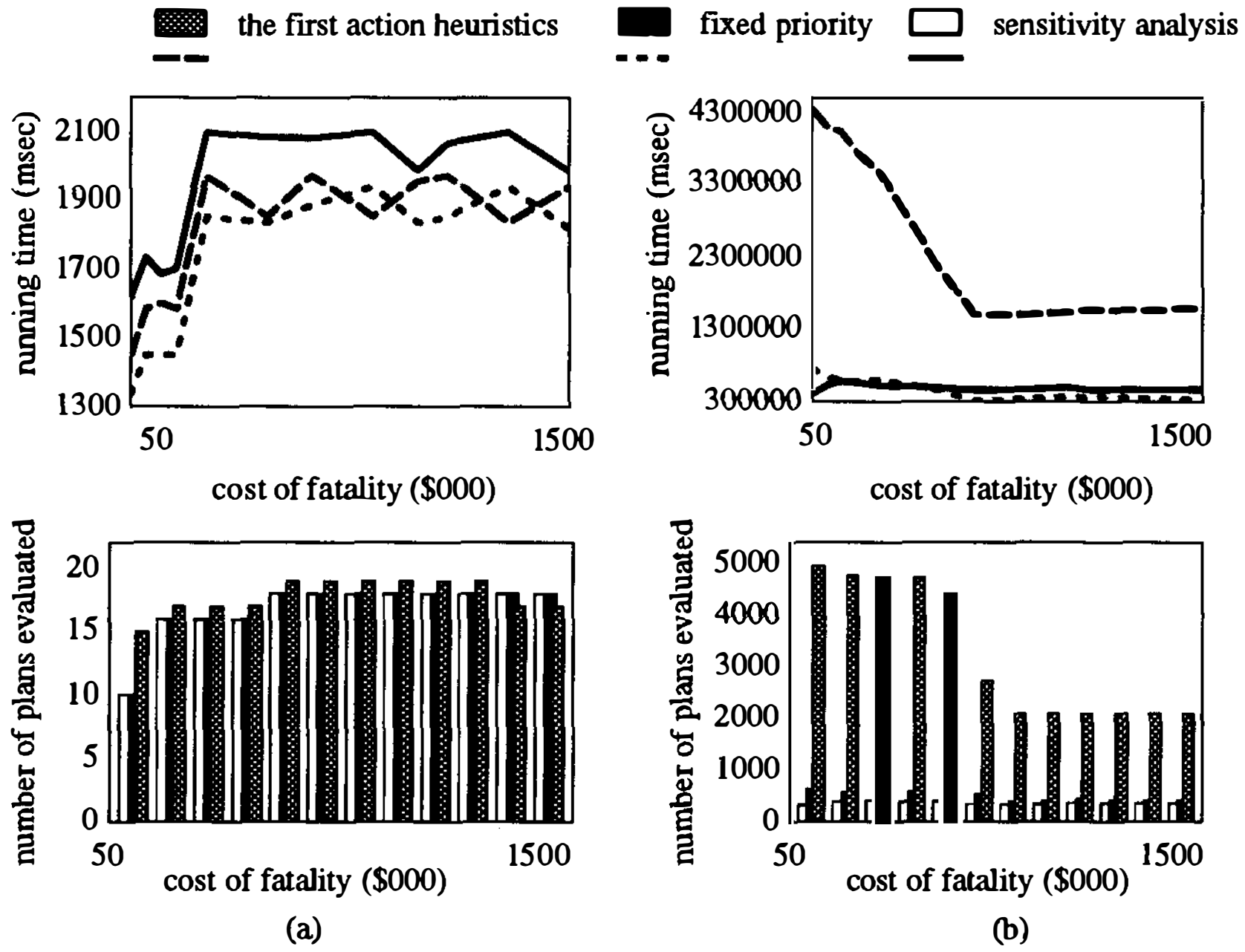

Figure 5: Results showing running time and number of plans evaluated versus the cost of fatality (a) for a small DVT domain (b) for a DVT domain of 6,206 plans.

We have successfully applied this technique to a variety of domains to reduce the number of plans before applying the DRIPS algorithm. For example, in the DVT domain the number is reduced from 6,206 to 232. Further results and detail will be reported in a future paper.

## References

[1] C. Boutilier and R. Dearden. Using abstractions for decision-theoretic planning with time constraints. In *Proceedings of the Twelfth National Conference on Artificial Intelligence*, pages 1016–1022, Seattle, July 1994.

[2] T. Dean, L. Pack Kaelbling, J. Kirman, and A. Nicholson. Planning with deadlines in stochastic domains. In *Proceedings of the Eleventh National Conference on Artificial Intelligence*, pages 574–579, July 1993.

[3] A. Doan and P. Haddawy. Generating macro operators for decision-theoretic planning. In *Working Notes of the AAAI Spring Symposium on Extending Theories of Action*, Stanford, March 1995.

[4] A.H. Doan and P. Haddawy. Decision-theoretic refinement planning: Principles and application. Technical Report TR-95-01-01, Dept. of Elect. Eng. & Computer Science, University of Wisconsin-Milwaukee, January 1995. Available via anonymous FTP from pub/tech_reports at ftp.cs.uwm.edu.

[5] M.G. Finigan. Knowledge acquisition for decision-theoretic planning. In *Proceedings MAICSS'95*, pages 98–102, Carbondale, IL, April 1995.

[6] R.P. Goldman and M.S. Boddy. Epsilon-safe planning. In *Proceedings of the Tenth Conference on Uncertainty in Artificial Intelligence*, pages 253–261, Seattle, July 1994.

[7] P. Haddawy and S. Hanks. Utility models for goal-directed decision-theoretic planners. Technical Report 93-06-04, Department of Computer Science and Engineering, University of Washington, June 1993. Available via anonymous FTP from ~ftp/pub/ai/ at cs.washington.edu.

[8] Hillner BE, Philbrick JT, Becker DM. Optimal management of suspected lower-extremity deep vein thrombosis: an evaluation with cost assessment of 24 management strategies. *Arch Intern Med*, 152:165–175, 1992.

[9] CE Kahn, Jr and P Haddawy. Management of suspected lower-extremity deep venous thrombosis (letter). *Archives of Internal Medicine*, 155:426, February 1995.

[10] N. Kushmerick, S. Hanks, and D. Weld. An algorithm for probabilistic least-commitment planning. In *Proceedings of the Twelfth National Conference on Artificial Intelligence*, pages 1073–1078, Seattle, 1994.

[11] M.P. Wellman. *Formulation of Tradeoffs in Planning Under Uncertainty*. Pitman, London,UK, 1990.

[12] G. A. Whitmore and M. C. Findlay. *Stochastic Dominance: An Approach to Decision Making Under Risk*. D. C. Health and Company, Lexington, MA, 1978.

[13] M. Williamson and S. Hanks. Optimal planning with a goal-directed utility model. In *Proceedings of the Second International Conference on Artificial Intelligence Planning Systems*, pages 176–181, Chicago, June 1994.